\documentclass[11pt]{article}

\usepackage[preprint]{acl}

\usepackage{amsmath,amsfonts,bm}

\def\eqref#1{equation~\ref{#1}}

\def\1{\bm{1}}

\DeclareMathAlphabet{\mathsfit}{\encodingdefault}{\sfdefault}{m}{sl}
\SetMathAlphabet{\mathsfit}{bold}{\encodingdefault}{\sfdefault}{bx}{n}

\usepackage{amssymb}
\newcommand{\registerremarkcolor}[2]{%
    \expandafter\def\csname color@#1\endcsname{#2}%
}

\newcommand{\remark}[2]{%
    \textcolor{\csname color@#1\endcsname}{%
        $\blacktriangleright$ \textbf{#1}: #2 $\blacktriangleleft$%
    }%
}

\registerremarkcolor{YZY}{cyan}
\usepackage{amsthm}
\theoremstyle{definition}
\newtheorem{definition}{Definition}[section]
\usepackage{algorithm}
\usepackage{algpseudocode}
\usepackage{enumitem}

\usepackage{times}
\usepackage{latexsym}

\usepackage[T1]{fontenc}

\usepackage[utf8]{inputenc}

\usepackage{microtype}

\usepackage{inconsolata}

\usepackage{graphicx}
\usepackage{CJKutf8}
\title{WASD: Locating Critical Neurons as Sufficient Conditions for\\
Explaining and Controlling LLM Behavior}

\author{
 \textbf{Haonan Yu\textsuperscript{1}},
 \textbf{Junhao Liu\textsuperscript{2}},
 \textbf{Zhenyu Yan\textsuperscript{3}},
 \textbf{Haoran Lin\textsuperscript{4}},
\\
 \textbf{Xin Zhang* \textsuperscript{5}},
\\
\\
 Key Lab of High Confidence Software Technologies (Peking University), Ministry of Education
 \\
 School of Computer Science, Peking University, Beijing, China
\\
    \{haonanyu\textsuperscript{1},
    zhenyuyan\textsuperscript{3},
    haoranlin\textsuperscript{4}\}@stu.pku.edu.cn,
    \{liujunhao\textsuperscript{2},
    xin\textsuperscript{5}\}@pku.edu.cn
}

\begin{document}
\maketitle

\begin{abstract}

Precise behavioral control of large language models (LLMs) is critical for complex applications. 
However, existing methods often incur high training costs, lack natural language controllability, or compromise semantic coherence.
To bridge this gap, we propose WASD (unWeaving Actionable Sufficient Directives), a novel framework that explains model behavior by identifying sufficient neural conditions for token generation. Our method represents candidate conditions as neuron-activation predicates and iteratively searches for a minimal set that guarantees the current output under input perturbations. 
Experiments on SST-2 and CounterFact with the Gemma-2-2B model demonstrate that our approach produces explanations that are more stable, accurate, and concise than conventional attribution graphs. Moreover, through a case study on controlling cross-lingual output generation, we validated the practical effectiveness of WASD in controlling model behavior.

\end{abstract}
\section{Introduction}

As large language models (LLMs) become integral to complex applications, there is a critical demand for precise behavioral control. Existing black-box approaches, such as alignment fine-tuning, struggle to guarantee reliable outcomes under novel conditions. On the other hand, mechanistic interpretability~\cite{naseem2026mechanistic,kowalska2025unboxing}, particularly circuit analysis, has recently gained significant popularity as a paradigm for reverse-engineering the internal mechanisms of neural networks. Among existing approaches, Circuit Tracer~\cite{ameisen2025circuit} represents the state of the art, identifying computational pathways by tracing the linear contributions of individual components to a final output token.

However, these methods are attribution-based, which face significant limitations. High attribution guarantees neither necessity nor sufficiency: components with massive linear contributions are not inherently causally required for the target token. Furthermore, these dense graphs are highly unstable under minor input perturbations and lack the clarity of rule-based explanations~\cite{Ribeiro_2018}.

\begin{figure*}[t] 
  
  \centering
  \includegraphics[width=2.0\columnwidth]{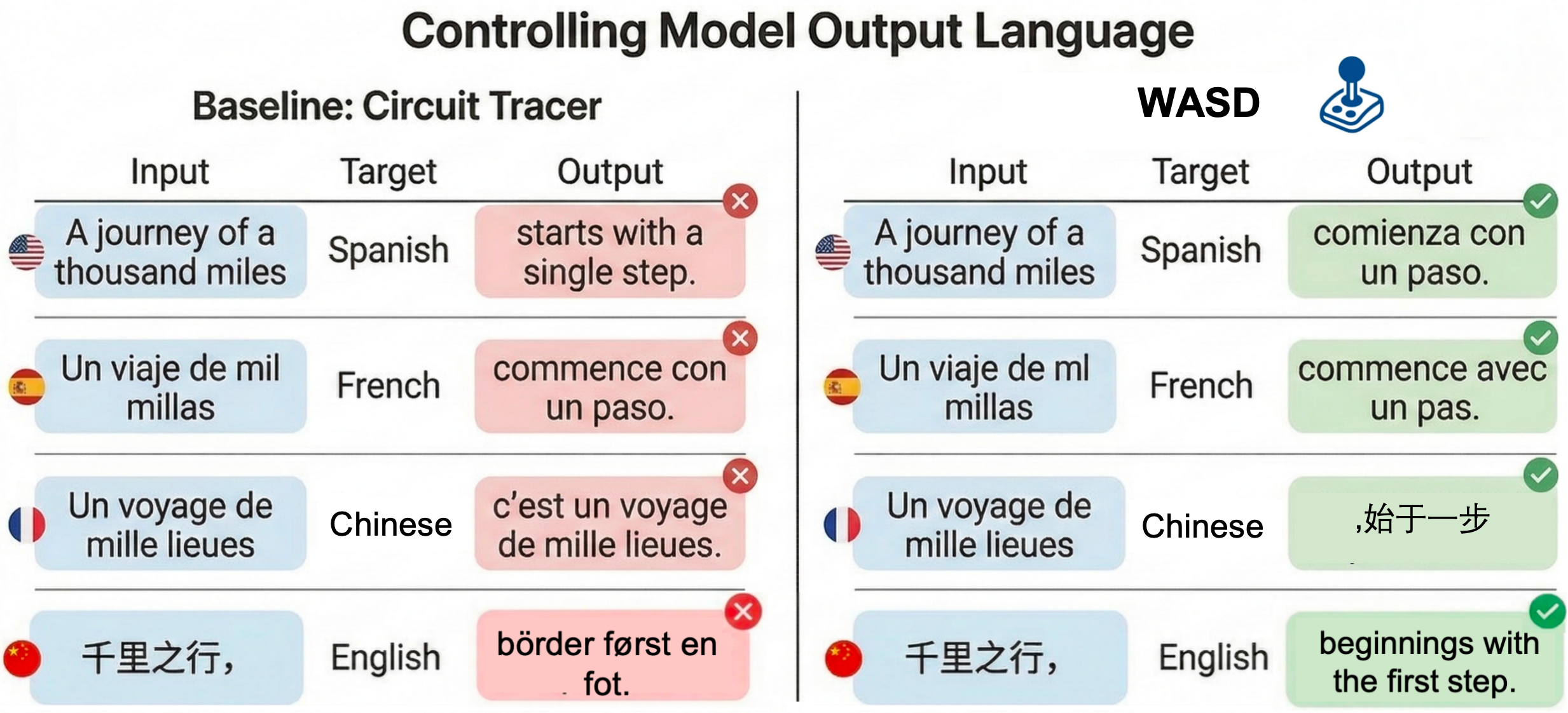}
  \caption {Comparison of language control capabilities between WASD and Circuit Tracer. In each subplot, the left panel displays the input, and the right panel displays the model's output following intervention. WASD controls the output by fixing specific neurons associated with the target language. In contrast, Circuit Tracer fixes the five highest-scoring neurons derived from the target language prompt's contribution map. 
  }
  \label{fig:comparison}
\end{figure*}

Figure~\ref{fig:comparison} (left) illustrates these shortcomings, demonstrating that using Circuit Tracer to control output language is unreliable. For example, when intervening on the top-five neurons in the target language's attribution graph to complete a phrase, this baseline frequently fails, generating either the original input language or incoherent text.

To resolve these limitations and enable the precise manipulation of LLMs,
we propose a framework named unWeaving Actionable Sufficient Directives from tangled circuit graphs (WASD).
Rather than relying on linear contributions, WASD identifies a minimal set of neural predicates—each defined by a neuron and its activation constraint—whose satisfaction is sufficient to maintain the target output token. To avoid the prohibitive computational cost of a direct search, the attribution scores from Circuit Tracer are leveraged as a heuristic. By ranking neurons according to their contribution, a greedy search procedure is guided to iteratively incorporate predicates into candidate explanations, efficiently extracting sufficient conditions.

Returning to the example, Figure~\ref{fig:comparison} (right) shows how WASD successfully enforces the target language constraint. By intervening directly on the identified neural predicates, the model consistently and accurately generates text in the specified language, regardless of the input.

WASD is evaluated on the SST-2 and CounterFact datasets using the Gemma-2-2B model. Results demonstrate that the produced explanations are significantly more stable, accurate, and concise than traditional attribution graphs. Furthermore, the framework is extended to enforce broader behavioral conditioning. By identifying and fixing neurons that guarantee a specific constraint—such as Chinese generation—the model consistently produces the desired output across diverse cross-lingual inputs.

In summary, the primary contributions of this work are threefold:

\begin{itemize}[nosep]
    \item The proposal of WASD, a novel mechanistic interpretability framework that explains model behavior by extracting sufficient neural conditions for token generation.
    \item Verification through experiments on multiple datasets, demonstrating that the sufficient conditions identified by WASD outperform Circuit Tracer in terms of accuracy, stability, and simplicity.
    \item A case study showing that the neurons identified by WASD can be used to control model outputs, such as ensuring consistent cross-lingual text generation.
\end{itemize}

\section{Background}

This section provides the necessary background knowledge that is integral to understanding our approach. We first briefly introduce the architecture of Transformer and the Circuit Tracer, then give a formal definition of the output of Circuit Tracer, and finally give a formal definition of the important components introduced by our method.

\subsection{Architecture of Transformer}

Transformer-based architectures~\cite{vaswani2017attention} form the foundation of modern large language models. A Transformer processes an input sequence of tokens by mapping each token to a vector embedding and passing these representations through a stack of layers. Each layer typically consists of a multi-head self-attention module followed by a feed-forward network (often implemented as a multi-layer perceptron, or MLP), with residual connections and layer normalization applied between components.

During inference, these embeddings propagate through the layers. The final hidden representation is projected into a logit vector over the vocabulary, and a softmax function determines the next token's probability distribution. Understanding how intermediate neurons and attention mechanisms contribute to these final logits is a core challenge in mechanistic interpretability.

\subsection{Circuit Tracer and Attribution Graphs}

A recent approach to mechanistic interpretability is Circuit Tracer, proposed by Anthropic, which aims to reveal the computational pathways responsible for specific model outputs.

\textbf{Replacement Model}. Circuit Tracer constructs an interpretable replacement model that approximates the behavior of a Transformer while exposing internal computational dependencies. In this approach, parts of the original model are replaced with more interpretable components that are trained to approximate the original computation. This replacement model enables tracing how intermediate activations influence one another and ultimately affect the output. Using this model, Circuit Tracer decomposes the computation into a set of interactions between interpretable units, enabling the construction of a linear attribution graph describing how information flows through the network during inference on a specific prompt. Importantly, the replacement model makes the features and neurons of the model more correlated, meaning that a single neuron often only represents a single feature.
As a result, it becomes possible to analyze the model's overall behavior by examining just a small fraction of neurons.

\textbf{Attribution Graph.} The output of Circuit Tracer is an attribution graph, which provides a structured representation of the computation performed by the model on a given prompt. Formally, the attribution graph can be represented as a directed weighted graph

$$
G=(V,E)
$$

where \(V\) denotes a set of nodes corresponding to neurons (or other computational units, such as input and output tokens) in the model. Each neuron \(v \in V\) has an associated activation value \(a_v\). \(E\) denotes directed edges representing influence relationships between neurons.

Each edge \( (u, v) \in E \) has an associated contribution weight \(c_{u \rightarrow v}\), which measures how the activation of neuron (\(u\)) contributes to the activation of neuron (\(v\)). In addition, neurons may have direct contributions to the output logits, representing their influence on the probability of generating specific tokens. This graph captures the causal structure of computation for a specific input prompt, allowing researchers to identify important neurons and pathways responsible for particular model behaviors.

\subsection{Sufficient Conditions for Model Behaviors}

To analyze and control model behaviors, we introduce the following formal definitions.

Let $\mathcal{V}$ denote a finite vocabulary of tokens.

\begin{definition}[Prompt]
    A prompt $x$ is a finite sequence of tokens $x=(t_1, t_2, \dots, t_n) \in \mathcal{V}^n$, where $n \in \mathbb{N}$. The space of all possible prompts is denoted as $\mathcal{X}=\bigcup_{n \in \mathbb{N}} \mathcal{V}^n$.
\end{definition}

Let $f: \mathcal{X} \to \mathcal{V}$ represent a language model. 
Let $\mathcal{H}$ be the set of all neurons in $f$, and let $\Phi_v(x) \in \mathbb{R}$ denote the activation value of a specific neuron $v \in \mathcal{H}$ during the forward pass of prompt $x$. Furthermore, let $C_v(x) \in \mathbb{R}$ denote the contribution of neuron $v$ to the logits of the output token.

Since our method only requires the neuron activation values and their contributions to the output token obtained by the Circuit Tracer, we formally define it as a function mapping a prompt to these specific internal states:

\begin{definition}[Weight Extraction]

The weight extraction function $WtExt$ takes a prompt $x \in \mathcal{X}$ as input and outputs a set containing the activation value and the corresponding logit contribution for each neuron:$$WtExt(x) = \{(\Phi_v(x), C_v(x)) \mid v \in \mathcal{H}\}$$

\end{definition}

\begin{definition}[Predicate]

A predicate $p$ is a tuple $(v, a) \in \mathcal{H} \times \mathbb{R}$, denoted symbolically as $p_{v,a}$, representing the logical constraint $\Phi_v(x)=a$.
    
\end{definition}

\begin{definition}[Rule]
A rule $r$ is a finite set of predicates defining a logical conjunction. For an index set $K=\{1, \dots, k\}$, a rule is defined as:

$$r:=\bigwedge_{i \in K} p_{v_i, a_i}$$

\end{definition}

Let $\Omega$ denote a set of discrete edit operations (deletion or replacement).

\begin{definition}[Neighborhood]

For a given prompt $x \in \mathcal{X}$, the neighborhood $\mathcal{N}(x)$ is defined as:
$$\mathcal{N}(x)=\{x' \in \mathcal{X} \mid x' \text{ is derived from } x \text{ via } \Omega\}$$

\end{definition}

Let $do(r)$ denote intervention, which hard-codes the internal model state such that $\forall p_{v_i, a_i} \in r, \Phi_{v_i}(\cdot)=a_i$. Let $f(\cdot \mid do(r))$ represent the model function operating under this fixed intervention. Let $\mathcal{D}_{\mathcal{N}}$ be a probability measure over $\mathcal{N}_\epsilon(x)$.

\begin{definition}[Precision]
\label{def:prec}

The precision of a rule $r$ with respect to a baseline prompt $x$, denoted $\text{Prec}(r, x)$, is the probability of output invariance under the intervention $do(r)$ across the local neighborhood:

$$\text{Prec}(r, x)=\mathbb{P}_{x' \sim \mathcal{D}_{\mathcal{N}}}\left[f(x' \mid do(r)) = f(x)\right]$$

\end{definition}

\begin{definition}[Sufficiency]

Given a threshold $\tau \in (0, 1]$, a rule $r$ constitutes a sufficient condition for the model's behavior on $x$ if:
$\text{Prec}(r, x) \geq \tau$

\end{definition}

Our goal is to identify a minimal set of jointly active neurons to maintain a decision, which can be formalized as the following optimization problem:

$$\arg\min_{r} |r| \quad \text{subject to} \quad \text{Prec}(r, x) \geq \tau$$

\section{Methodology}

\begin{figure*}[t] 
  
  \centering
  \includegraphics[width=2.0\columnwidth]{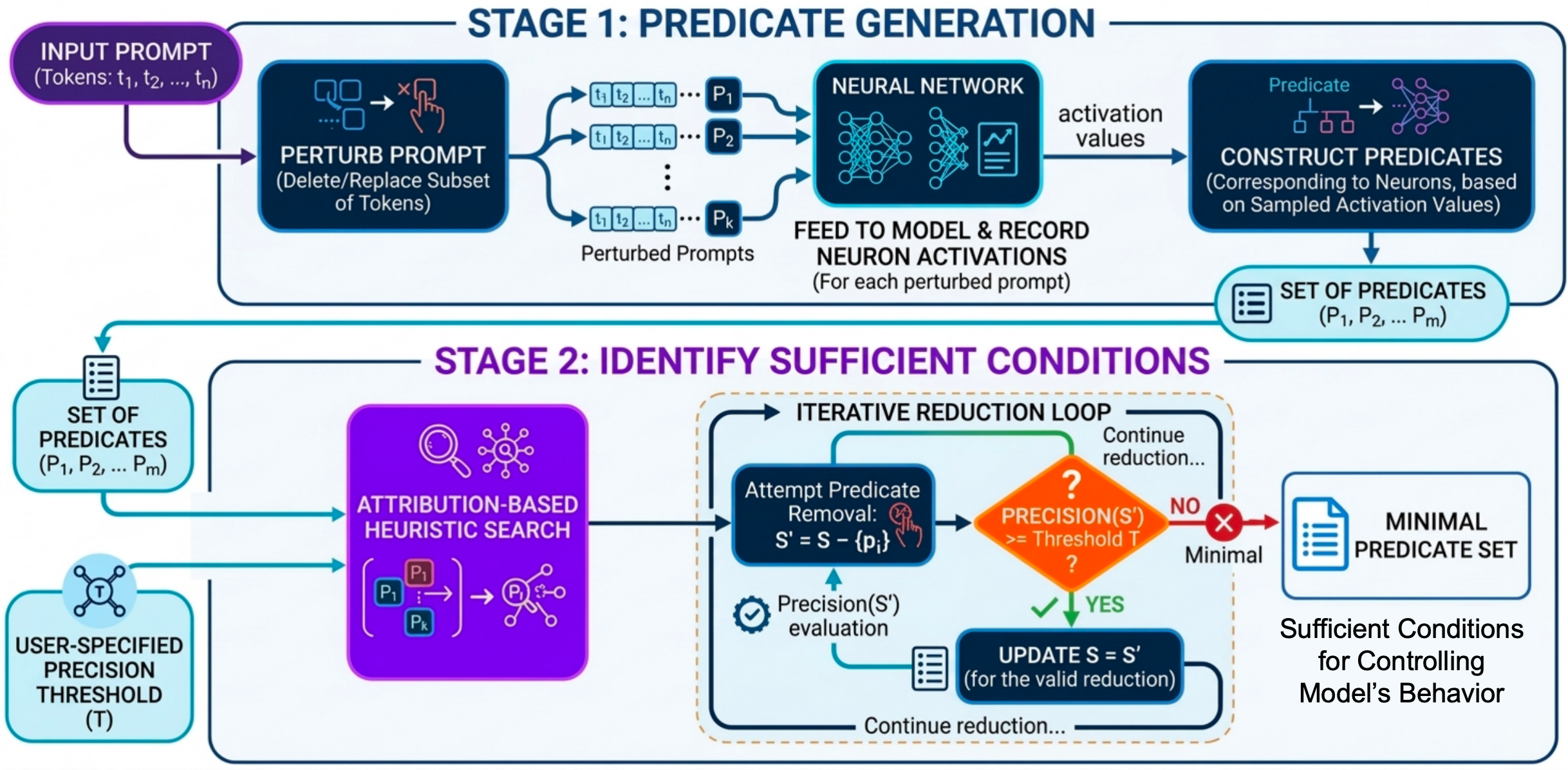}
  \caption {The workflow of WASD.}
  \label{fig:workflow}
\end{figure*}

This section outlines WASD, our proposed method for extracting the sufficient conditions that govern model behavior. As illustrated in Figure~\ref{fig:workflow}, the methodology operates in two primary phases: generating candidate predicates via prompt perturbation and activation tracking, followed by an attribution-guided heuristic search and iterative reduction loop. Together, these steps construct a minimized rule set that is both concise and sufficient to maintain model precision.

\subsection{Generating Predicates}

\begin{algorithm}[t]
   \caption{Generating Predicates}
   \label{alg:GenP}
    \textbf{Input}: Original prompt $x=( t_1,t_2,...,t_n)$, the model to explain $f$\\
    \textbf{Parameter}: The scaling factor $\lambda$\\
    \textbf{Output}: Set of candidate predicates with contributions $\mathbb{P}$
    
    \begin{algorithmic}[1]
        \State $\mathcal{N}(x)$ = Perturbation(x)
        \State Initialize activation map A = \{\}
        \State Initialize contribution map C = \{\}
        \State \# Calculate activations and contributions
        \For {$x' \in \mathcal{N}(x)$}
            \State Activations, Contributions = WtExt(x')
            \For {Each neuron $n$}
                \State A[n].add(Activations[n])
                \State C[n].add(Contributions[n])
            \EndFor
        \EndFor
        \State Initialize predicates $\mathbb{P}$ = \{\}
        \State \# Formulate predicates
        \For {Each neuron $n$}
            \State $A_{max, n} = MAX(A[n])$
            \State $p_n = A_{max, n} * \lambda$
            \State $c_n = MEAN(C[n])$
            \State $Add (p_n, c_n)\ to\ \mathbb{P}$
        \EndFor
        \State Return $\mathbb{P}$
    \end{algorithmic}
\end{algorithm}

Algorithm~\ref{alg:GenP} details the predicate generation process. Inspired by traditional perturbation-based interpretation techniques like LIME~\cite{Ribeiro_2016} and Anchors~\cite{Ribeiro_2018}, a local neighborhood is first constructed by editing the input prompt (Line 1), and maps are initialized to track activations and logit contributions (Lines 2–3). For each perturbed prompt, Circuit Tracer computes the attribution graph (Line 6), recording individual neuron activations and contributions across varying inputs (Lines 8–9). From these statistics, candidate predicates are constructed. Each neuron's maximum recorded activation is multiplied by a scaling factor $\lambda$ to define its predicate threshold. This predicate, alongside the neuron’s average contribution, is then stored in the candidate predicate set $\mathbb{P}$ (Lines 14–19).

Based on these statistics, candidate predicates are constructed. For each neuron, our method identifies its maximum recorded activation and multiply it by a scaling factor $\lambda$ to define the predicate threshold. The predicate, together with the neuron’s average contribution, is then stored in the candidate predicate set $\mathbb{P}$ (Lines 14–19).

The hyperparameter $\lambda$ dictates the granularity of the resulting explanations. However, a larger $\lambda$ does not strictly yield better performance; an excessively high value can disproportionately amplify a single neuron's importance, obscuring the synergistic network interactions that typically drive the model's final prediction. To prevent the method from degrading into an analysis of a single neuron's preferred output, a grid search strategy is applied to empirically determine the optimal $\lambda$ that maximizes precision.

\subsection{Identifying Sufficient Conditions}

\begin{algorithm}[t]
   \caption{Identifying Sufficient Conditions}
   \label{alg:CalcRule}
    \textbf{Input}: A prompt $x$, the candidate predicates $\mathbb{P}$\\
    \textbf{Parameter}: The precision threshold $\tau$\\
    \textbf{Output}: The minimal sufficient rule set $r$
    
    \begin{algorithmic}[1]
        \State Sort $\mathbb{P}$ by contribution in descending order.
        \State Initialize rule $r = \emptyset$
        \State Current\_Prec$ = prec(\emptyset, x)$
        \State \# Additive Search
        \For {$(p,c) \in \mathbb{P}$}
            \State $r_{temp} = r\cup p$
            \State New\_Prec$ = prec(r_{temp}, x)$
            \If {New\_Prec $>$ Current\_Prec}
                \State $r = r_{temp}$
                \State Current\_Prec = New\_Prec
            \EndIf
            \If {Current\_Prec$ \geq \tau$}
                \State BREAK
            \EndIf
        \EndFor
        \State \# Pruning
        \For {$p \in r$}
            \State $r_{temp}=r\ -\{p\}$
            \If{$prec(r_{temp}, x) \geq \tau$}
                \State $r=r_{temp}$
            \EndIf
        \EndFor
        \State Return $r$
    \end{algorithmic}
\end{algorithm}

Once the candidate predicates are generated, Algorithm~\ref{alg:CalcRule} employs a contribution-based heuristic search to identify the sufficient conditions controlling the model’s output. The generated predicates are first sorted by descending contribution (Line 1). Next, an empty candidate rule set is initialized, and the baseline precision, $prec(\emptyset, x)$, is calculated within the prompt's neighborhood without any interventions (Lines 2–3).
The rule set is then iteratively expanded by introducing new predicates (Lines 5–6). A predicate is retained only if it improves the precision of the rule set; otherwise, it is discarded (Lines 7–11). This additive phase continues until the current rule set's precision, $prec(r, x)$, meets or exceeds a predefined threshold $\tau$ (Lines 12–14). Because the candidate pool encompasses neurons representing all input tokens, the search is guaranteed to converge on a rule that satisfies this precision constraint.
Finally, to ensure conciseness, a pruning step is performed. The finalized rule set is systematically evaluated to drop one predicate at a time (Line 18). If removing a predicate does not cause the precision to fall below $\tau$, it is deleted (Lines 19–21). This strips the rule set of any redundant predicates, ultimately returning a minimized rule that serves as the sufficient condition for controlling the model's behavior (Line 23).
\section{Experiment}

In this section, we empirically evaluate the effectiveness of our proposed neuron identification method. The experiments are designed to answer two key questions: (1) whether the neurons identified by our method genuinely govern the model's behavior, and (2) whether our approach provides more stable and concise explanations compared with existing attribution-based methods.

\subsection{Experimental Setup}

To evaluate the generality of the proposed method, experiments are conducted on two distinct tasks. The first task involves sentiment classification utilizing the SST-2 dataset~\cite{socher-etal-2013-recursive}, where its test set comprising 1,820 sentences is used to generate explanations. The second task focuses on text completion using the CounterFact dataset~\cite{meng2022locating}, specifically employing its paraphrase\_prompts subset of 2,191 sentences as a test set for explanation generation. All experiments are performed on the Gemma-2-2B model~\cite{gemma_2024}.

Across all tasks, the parameters of our method are set to $\tau = 0.9$ and $\lambda = 6.5$. These values are determined through grid search to achieve a good balance between explanation size and fidelity. As a baseline, we compare our approach with the original Circuit Tracer method. To ensure a fair comparison and isolate the effect of neurons' activations, we fix the neurons identified by Circuit Tracer (Top-3, Top-5, and Top-10 contributors) to their original activation values multiplied by coefficients $\lambda_3$, $\lambda_5$, and $\lambda_{10}$, respectively. These coefficients were also optimized via grid search to yield the highest average precision for the baseline. This setup allows us to demonstrate that the model’s behavior is governed by the specific set of neurons we identify rather than merely the magnitude of activation values.

The experimental procedure is as follows. First, we feed text prompts into the model and analyze the next-token prediction. Using both our method and the baseline, we identify the neurons that govern the model’s output along with their activation values. We then perturb the input prompts (e.g., by adding neutral prefixes, detailed in Chapter~\ref{neutralprefix} ) while fixing the identified neurons to their original activation states. Finally, we observe whether the model’s output remains consistent despite the external perturbations.

\subsection{Evaluation Metrics}

To quantify the performance of our method, we employ the following three metrics:

\paragraph{Precision (Fidelity).} As defined in Definition~\ref{def:prec}, Precision measures the fidelity of an explanation. Specifically, it is defined as the probability that the model’s output remains unchanged under perturbed conditions when the identified neurons are fixed to a specific activation values.

\paragraph{Instability.} Instability measures the consistency of explanations under small input perturbation. We prepend neutral prefixes that do not alter the ground-truth output. Let $r$ and $r'$ denote the neuron sets identified for the original prompt and the prefix-augmented prompt, respectively. Instability is defined using the Jaccard distance between these sets:
\[
Instability = \frac{|r \cup r'| - |r \cap r'|}{|r \cup r'|}
\]
A smaller Jaccard distance indicates that the explanation remains more similar after perturbation, resulting in greater stability.

\paragraph{Size.} This simply measures the number of neurons required to form the explanation. A smaller size indicates a more concise and interpretable explanation of the model's internal logic.

\subsection{Experiment Results and Analysis}

The primary results of our comparative analysis are summarized in Table~\ref{tab:results_comparison}.

\begin{table*}[]
    \centering
    \begin{tabular}{ccccc}
    \hline
          Method & Task & Precision ↑ & Instability ↓ & Size (Neurons) ↓ \\        
    \hline
         Circuit Tracer (Top-3) & SST2 & 46.37\%(±1.45\%) & 0.166(±0.109) & 3\\   
         Circuit Tracer (Top-5) & SST2 & 40.88\%(±1.36\%) & 0.190(±0.076) & 5\\   
         Circuit Tracer (Top-10) & SST2 & 30.95\%(±1.35\%) & 0.052(±0.037) & 10\\
         WASD & SST2 & \textbf{91.92\%(±1.34\%)} & \textbf{0.048(±0.056)} & 1.53(±0.11) \\
    \hline
         Circuit Tracer (Top-3) & CounterFact & 48.04\%(±3.28\%) & 0.212(±0.041) & 3\\      
         Circuit Tracer (Top-5) & CounterFact & 48.69\%(±3.30\%) & 0.250(±0.035) & 5\\   
         Circuit Tracer (Top-10) & CounterFact & 50.33\%(±3.41\%) & 0.260(±0.029) & 10\\
         WASD & CounterFact & \textbf{92.60\%(±0.69\%)} & \textbf{0.103(±0.032)} & 2.82(±0.27)\\
    \hline
    \end{tabular}
    \caption{Performance comparison between WASD and Circuit Tracer on SST2 and CounterFact datasets (mean ± 95\% confidence interval).}
    \label{tab:results_comparison}
\end{table*}

As indicated by the experimental data, our proposed method consistently achieves higher perturbation accuracy and superior stability compared to the Circuit Tracer baselines across both datasets.

Specifically, in the sentiment classification task (SST2), our method identified a more compact set of neurons that maintained model output consistency even under significant prompt perturbation. In terms of stability, the Jaccard similarity for our identified neurons remained significantly higher than the Top-K neurons selected by Circuit Tracer when neutral prefixes were introduced. This suggests that our method captures the functional core of the model's decision-making process rather than just transient activations.

Furthermore, our method achieves these gains while maintaining a smaller Size—often requiring fewer neurons than the Top-10 or even Top-5 baselines to reach higher fidelity. This confirms that our approach effectively filters out noise and isolates the specific neurons that govern the model's behavior, providing a more concise and reliable explanation of LLM internals.

\section{Case Study}

To further demonstrate the efficacy of our approach, we introduce a feature within WASD that allows users to set custom constraints for the model's output. Specifically, this feature identifies the neuron-level sufficient conditions required to guarantee that the output belongs to a designated target set.

Technically, we integrate a Llama-3-7B-Instruct~\cite{Llama} model into the WASD pipeline as an automated evaluator to determine whether the target model's output satisfies the user-defined constraints. Based on this evaluation, the framework locates the precise sufficient conditions needed to enforce arbitrary behavioral rules. In this section, we use cross-lingual generation—specifically, forcing the model to output Chinese when prompted in English—as a case study to validate our method's capacity for fundamental behavior control.

\subsection{Experimental Setup}

We define the target user constraint as: "Ensure the model's output remains in Chinese." We use Gemma-2-2B as the target model for both interpretation and intervention. The process consists of two steps:
The process contains two steps:

\textbf{Condition Extraction}: We input several Chinese prompts alongside the specified constraint into the WASD framework. The framework then identifies the sufficient conditions—a specific subset of neurons and their activation values—that govern Chinese text generation. The neurons utilized by both WASD and our Circuit Tracer baseline were derived from analyzing these same Chinese prompts (Appendix\ref{ChinesePrompts}).

\textbf{Neuron Intervention}: We utilize the CounterFact dataset, which consists entirely of English text sequences. During the text completion phase, we intervene in Gemma-2-2B by fixing the activation states of the previously identified neurons to their target values, testing the hypothesis that this localized intervention is sufficient to force the model to complete the English prompts in Chinese.

\subsection{Evaluation Metrics and Baseline}
To verify that the identified neurons specifically govern language generation without degrading the model's fundamental linguistic capabilities or instruction-following semantics, we benchmarked our WASD-based intervention against two baselines: standard In-Context Learning~\cite{brown2020language} (ICL) and Circuit Tracer. The evaluation spans two dimensions:

\textbf{Control Efficacy}: We employ FastText Language Identification~\cite{joulin2016bag} (LID) to determine the exact percentage of generated completions that successfully transition to Chinese.

\textbf{Quality and Semantic Coherence}: We assess whether the intervention damages the model's semantic representations. We calculate the LaBSE~\cite{Feng_2022} cosine similarity between the original English input and the generated Chinese output to measure cross-lingual semantic retention. Furthermore, we compute the Perplexity~\cite{jurafsky2024speech} (PPL) using a Qwen2.5-7B-Instruct~\cite{qwen2.5} model to evaluate the fluency of the generated text.

\subsection{Results and Analysis}
\begin{table*}[]
    \centering
    \begin{tabular}{ccccc}
    \hline
          Method & FastText LID (\%) ↑ & LaBSE Similarity ↑ & Perplexity (PPL) ↓ \\        
    \hline
         Circuit Tracer & 14.8\% & 0.2175(±0.032) & 331.84(±113.54) \\  
         In-Context Learning  & 60.4\% & 0.2435(±0.032) & 1162.26(±1310.1) \\   
         WASD & \textbf{89.4\%} & \textbf{0.2867(±0.023)} & \textbf{250.79(±54.45)} \\
    \hline
    \end{tabular}
    \caption{Performance comparison of cross-lingual output control on the CounterFact dataset (mean ± 95\% confidence interval).}
    \label{tab:case_study}
\end{table*}

As shown in Table~\ref{tab:case_study}, fixing the activation states of the neurons identified by our framework achieves a FastText LID score of 89.4\%, demonstrating control efficacy that significantly exceeds both explicit ICL prompts (60.4\%) and Circuit Tracer (14.8\%). 

Crucially, WASD achieves the highest LaBSE similarity at 0.2867, indicating that the model successfully retains the factual and semantic intent of the English CounterFact prompts despite the forced language switch. In contrast, Circuit Tracer yields the lowest semantic retention (0.2175). Finally, WASD maintains a low external PPL of 250.79. While Circuit Tracer also maintains a relatively low PPL (331.84), it ultimately fails at the primary language transition task. Conversely, ICL suffers from a massive PPL spike (1162.26), suggesting severe text degeneration. Overall, these results confirm that WASD accurately isolates the specific mechanism for language selection without disrupting the broader language modeling circuitry.
\section{Related Work}

Recent interpretability research seeks to understand and control the internal mechanisms of large neural networks. Two key directions are relevant here: mechanistic interpretability, which uncovers the computational structures driving model behavior, and intervention methods that leverage these insights to control outputs.

\subsection{Mechanistic Interpretability and Circuit Discovery}

Mechanistic interpretability reverse-engineers neural networks to find human-understandable algorithms embedded in their parameters~\cite{olah2020zoom,elhage2021mathematical}. A major focus is circuit discovery: identifying subnetworks responsible for specific behaviors. Techniques like causal tracing~\cite{meng2022locating}, path patching~\cite{wang2022interpretability}, and Automated Circuit Discovery (ACDC)~\cite{conmy2023towards} assess how intermediate representations affect final outputs. More recently, attribution-based methods like Circuit Tracer~\cite{ameisen2025circuit} quantify the linear contributions of components (e.g., attention heads or MLP layers) to predictions. However, these approaches primarily capture linear attribution rather than strict causality. As we demonstrate, highly attributed components are often neither necessary nor sufficient for a given behavior, highlighting the need for explanation frameworks grounded in stronger causal criteria.

\subsection{Controlling and Intervening on Model Behavior}

Mechanistic interpretability also enables precise control over model behavior. Currently, LLM outputs are typically guided indirectly via prompt engineering~\cite{brown2020language}, which can be fragile under adversarial attacks or context constraints. Alternatively, activation engineering and representation steering~\cite{zou2023representation,turner2024activation} intervene during the forward pass by injecting continuous activation vectors into intermediate layers. Recent feature steering methods, such as Goodfire's Auto Steer~\cite{sprejer2026mindperformancegapcapabilitybehavior}, automate this using natural-language queries. Yet, these interventions face a capability-behavior trade-off: they can severely degrade accuracy and coherence, sometimes underperforming simple prompting. Furthermore, they operate on high-dimensional representations rather than individual computational components. In contrast, our approach identifies minimal sets of causally sufficient neurons that govern specific behaviors. Anchoring these neurons enables targeted interventions—such as enforcing cross-lingual generation—without altering prompts or broadly shifting internal representations.
\section{Conclusion}

In this work, we introduced WASD, a novel mechanistic interpretability framework that explains language model behavior through the identification of sufficient neural conditions for token generation. Unlike traditional attribution-based approaches that primarily measure linear contribution, our framework formulates explanations as minimal sets of neuron-level predicates whose activation is sufficient to preserve the model’s output under input perturbations. By combining perturbation-based evaluation with attribution-guided heuristic search, WASD efficiently navigates the vast hypothesis space of possible neural states and extracts concise rule-based explanations.

Experimental results on SST-2 and CounterFact demonstrate that our approach produces explanations that are substantially more faithful, stable, and compact than those derived from conventional attribution graphs. In addition, our case study on cross-lingual output control shows that the neurons identified by WASD can be used to perform precise neuron-level interventions, enabling targeted manipulation of model behavior without relying on prompt engineering or broad activation steering.

Overall, our findings suggest that framing interpretability as the search for sufficient neural predicates provides a promising direction for bridging symbolic rule-based explanations and neural computation. In future work, we plan to extend WASD to larger language models and more complex behaviors, explore more efficient predicate search strategies, and investigate how neuron-level sufficient conditions relate to higher-level computational circuits within transformer architectures.
\newpage
\section{Limitations}

Our approach relies on a heuristic search guided by the neural contributions identified by the Circuit Tracer. Although our results demonstrate that this method effectively identifies neurons serving as sufficient conditions for the target outcome, the search strategy still leaves room for improvement. Specifically, because it is a heuristic approach, we cannot guarantee that the algorithm will consistently yield a globally optimal solution; instead, it may converge on a locally optimal one. Nevertheless, when balancing time costs against experimental outcomes, our method successfully achieves favorable results while maintaining a relatively low computational overhead.

\bibliography{custom}
\appendix
\label{sec:appendix}

\section{Experiments Compute Resources}
\label{sec:exp_compute_resources}

All the experiments were conducted on a server with 4 high-performance GPUs, each providing up to 10,000+ CUDA cores and 24GB of dedicated GPU memory, combined with 256GB system memory.

\section{Neutral Prefixes}
\label{neutralprefix}

In the stability experiments, we introduced prefixes sampled from a predefined set of neutral prefixes. These prefixes were added in a randomized order until the model's output remained unchanged relative to the output before the prefix was introduced. This procedure was designed to ensure, as much as possible, that the perturbations did not substantially alter the model's internal behavior.

The set of neutral prefixes used in our experiments is as follows:
\begin{itemize}
    \item ``As we all know, ''.
    \item ``Note that, ''.
    \item ``In fact, ''.
    \item ``Fact: ''.
    \item ``Text: ''.
    \item ``Input: ''.
\end{itemize}
\section{Chinese Prompts used in Case Study}
\label{ChinesePrompts}

In the Case Study experiments, WASD and Circuit Tracer were tasked with identifying—from the interpretation process of several Chinese prompts—the neurons that influence the Chinese output. The prompts utilized are as follows:


\begin{itemize}
    \item ``\begin{CJK*}{UTF8}{gbsn}上海所在国家的首都是\end{CJK*}'' (The capital of the country where Shanghai is located is)
    \item ``\begin{CJK*}{UTF8}{gbsn}千里之行，\end{CJK*}'' (A journey of a thousand miles,)
    \item ``\begin{CJK*}{UTF8}{gbsn}地球上水体面积最大的大洋是\end{CJK*}'' (The ocean with the largest body of water on Earth is)
    \item ``\begin{CJK*}{UTF8}{gbsn}苹果，香蕉和橘子都属于\end{CJK*}'' (Apples, bananas, and oranges all belong to)
\end{itemize}

\section{The Use of AI Assistants}

In this work, AI assistant was used solely for refining the language during the writing process and did not participate in any experimental, discussion, or coding aspects.

\section{The Computational Cost}

The time cost of our method can be primarily divided into two components: 1) calculating the attribution graph using Circuit Tracer, and 2) evaluating the precision of each candidate predicate through sampling.
In our experimental setup and with the available compute resources, calculating the attribution graph takes an average of 37.9 seconds. This part of the cost is also necessary for Circuit Tracer. And the additional cost of our approach, the sampling process, when utilizing 8 GPUs with equivalent computing resources for parallel processing, completes in an average of 43.85 seconds.
Although our approach incurs higher overhead, it provides a trade-off where increased computational resources result in higher precision and a more accurate identification of the sufficient conditions that control model behavior.

\section{Broader Impacts and Potential Risks}

While the WASD framework provides significant advancements in explaining and controlling large language model (LLM) behavior, it inherently presents dual-use risks.

The primary vulnerability stems from the framework's core capability: identifying a minimal set of neural predicates whose satisfaction is sufficient to maintain a target output. While this allows for beneficial interventions—such as ensuring consistent cross-lingual generation —it can also be exploited.

Specifically, if a target model possesses the latent capacity to generate offensive, biased, or harmful content, malicious actors could utilize WASD to isolate the exact neural conditions that govern the production of such speech. Because WASD allows users to intervene by hard-coding the internal model state to these specific activation values , an adversary could theoretically bypass standard prompt-level safety filters or alignment fine-tuning. By artificially fixing the activation states of these identified neurons, the model could be reliably forced to generate malicious outputs regardless of the input context.

Consequently, while our method advances precise behavioral control, it underscores the ongoing necessity for robust base-model safety and highlights the importance of utilizing mechanistic interpretability not just for control, but for identifying and scrubbing harmful latent circuits during the training phase.

\end{document}